\documentclass[english,letterpaper]{article}
\usepackage[T1]{fontenc}
\usepackage[utf8]{inputenc}
\usepackage{float}
\usepackage{amsmath}
\usepackage{amsthm}
\usepackage{graphicx}
\usepackage{microtype}

\makeatletter

\floatstyle{ruled}
\newfloat{algorithm}{tbp}{loa}
\providecommand{\algorithmname}{Algorithm}
\floatname{algorithm}{\protect\algorithmname}

\def\year{2019}\relax
\usepackage{aaai19}  
\usepackage{times}  
\usepackage{helvet}  
\usepackage{courier}  
\usepackage{url}  
\usepackage{graphicx}  
\usepackage{algorithm,algpseudocode}

\newcommand\blfootnote[1]{%
  \begingroup
  \renewcommand\thefootnote{}\footnote{#1}%
  \addtocounter{footnote}{-1}%
  \endgroup
}

\@ifundefined{showcaptionsetup}{}{%
 \PassOptionsToPackage{caption=false}{subfig}}
\usepackage{subfig}
\makeatother

\usepackage{babel}
\begin{document}

\title{Data Interpolations in Deep Generative Models\\
 under Non-Simply-Connected Manifold Topology}

\author{Jiseob Kim\\
Department of Computer Science\\
Seoul National University\\
1 Gwanak-ro, Gwanak-gu, Seoul 08826\\
\texttt{jkim@bi.snu.ac.kr}\\
 \And Byoung-Tak Zhang\\
Department of Computer Science\\
Seoul National University\\
1 Gwanak-ro, Gwanak-gu, Seoul 08826 \\
\texttt{btzhang@bi.snu.ac.kr}}
\maketitle
\begin{abstract}
Exploiting the deep generative model's remarkable ability of learning
the data-manifold structure, some recent researches proposed a geometric
data interpolation method based on the geodesic curves on the learned
data-manifold. However, this interpolation method often gives poor
results due to a topological difference between the model and the
dataset. The model defines a family of simply-connected manifolds,
whereas the dataset generally contains disconnected regions or holes
that make them non-simply-connected. To compensate this difference,
we propose a novel density regularizer that make the interpolation
path circumvent the holes denoted by low probability density. We confirm
that our method gives consistently better interpolation results from
the experiments with real-world image datasets. 

\end{abstract}

\section{Introduction}

Deep generative models such as Generative Adversarial Network (GAN)
\shortcite{goodfellow2014generative} and Variational Auto-Encoder (VAE)
\shortcite{kingma2013auto} show remarkable performance in learning the
manifold structure of the data. Among other advantages, knowledge
of the manifold structure allows interpolating the data points along
the manifold, which gives semantically plausible interpolation results.

Toward geometrically grounded interpolations on the manifold,
some recent works \shortcite{shao2017riemannian,chen2018metrics} proposed
a geodesic interpolation method. Unlike previous methods which naïvely
followed linear paths in the latent space, geodesic interpolation
methods find the shortest paths on the manifold based on the metric
induced from the ambient data-space. This shares the philosophy with
classic algorithms like Isomap \shortcite{tenenbaum2000global}, and well-grounded
on the Riemannian geometry. However, when experimented with real-world
data, it often gives poor results containing unrealistic interpolation
points. 

We argue that the problem stems from a topological mismatch between
the model and the data. Deep generative models define a simply-connected
manifold because the latent space is endowed with a probability distribution
of simply-connected support, and the smooth generative mapping\footnote{The smoothness of the generative mapping is a necessary condition
to use the geodesic interpolation methods.} preserves the latent-space topology in the data-space. On the other
hand, real-world datasets generally contains disconnected regions
or holes which make them non-simply-connected. As this topological
difference is a fundamental matter, deep generative models cannot
help including the holes in the dataset as valid regions in their
manifold representation. In consequence, the geodesic curve often
finds a shortcut that passes through the holes and the interpolation
points get unrealistic. 

To tackle this problem, we propose to add a density regularization
term to the path-energy loss. The idea comes from an observation that
even though deep generative models cannot be matched to the dataset
topology, the maximum-likelihood training forces the models to have
low probability densities where the holes exist. Therefore, if a low-density-penalizing
term is added to the path-energy loss, minimizing the loss will find
a path close to the geodesic while going around the holes denoted
by low densities. We demonstrate this method is effective and gives
semantically better interpolation results from the experiments.

\section{Geometric Interpolations with Density Regularizer}

\begin{figure}[t]
\begin{centering}
\includegraphics[width=0.85\columnwidth]{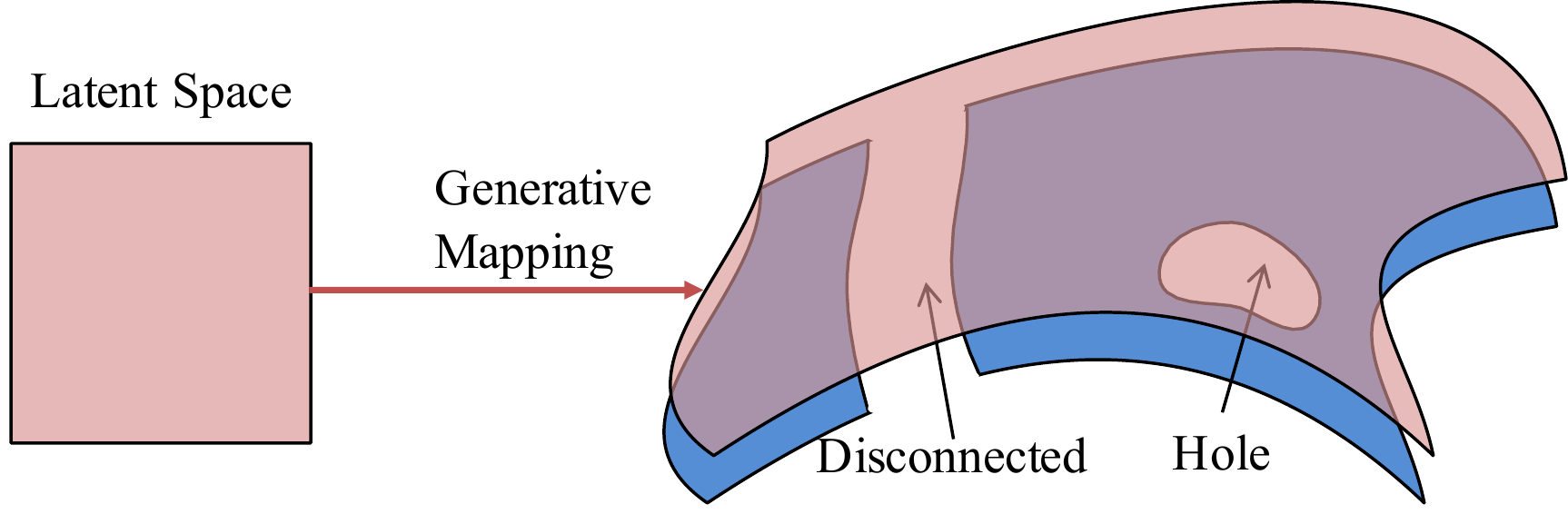}
\par\end{centering}
\caption{The true underlying structure of the data is generally non-simply-connected
(\emph{blue}), whereas the manifold induced by the deep generative
model is simply-connected (\emph{red}). This topological difference
degrades the quality of the interpolation computed on the deep generative
model. \label{fig:main_diagram}}
\end{figure}

\subsection{Notations}

We denote the latent space as $Z$, the input data space as $X$ and
the coordinates of each space as $z$ and $x$ respectively. We do
not use bold letter for the vectors or matrices, but brackets $[\cdot]$
are used when explicit indication of matrix is needed.

\begin{figure*}
\centering{}\subfloat[]{\includegraphics[width=0.2\textwidth]{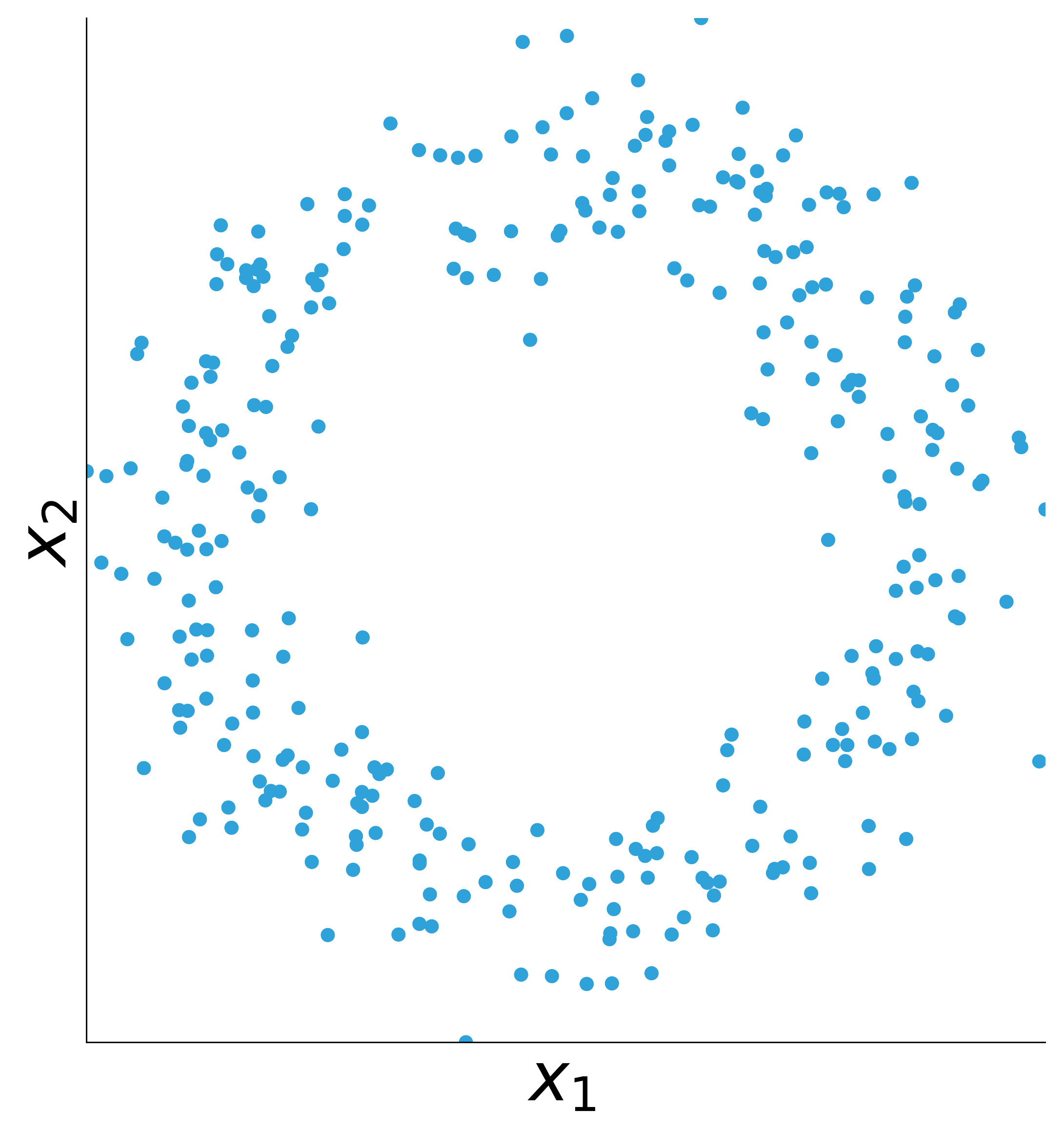}

}\subfloat[]{\includegraphics[width=0.2\textwidth]{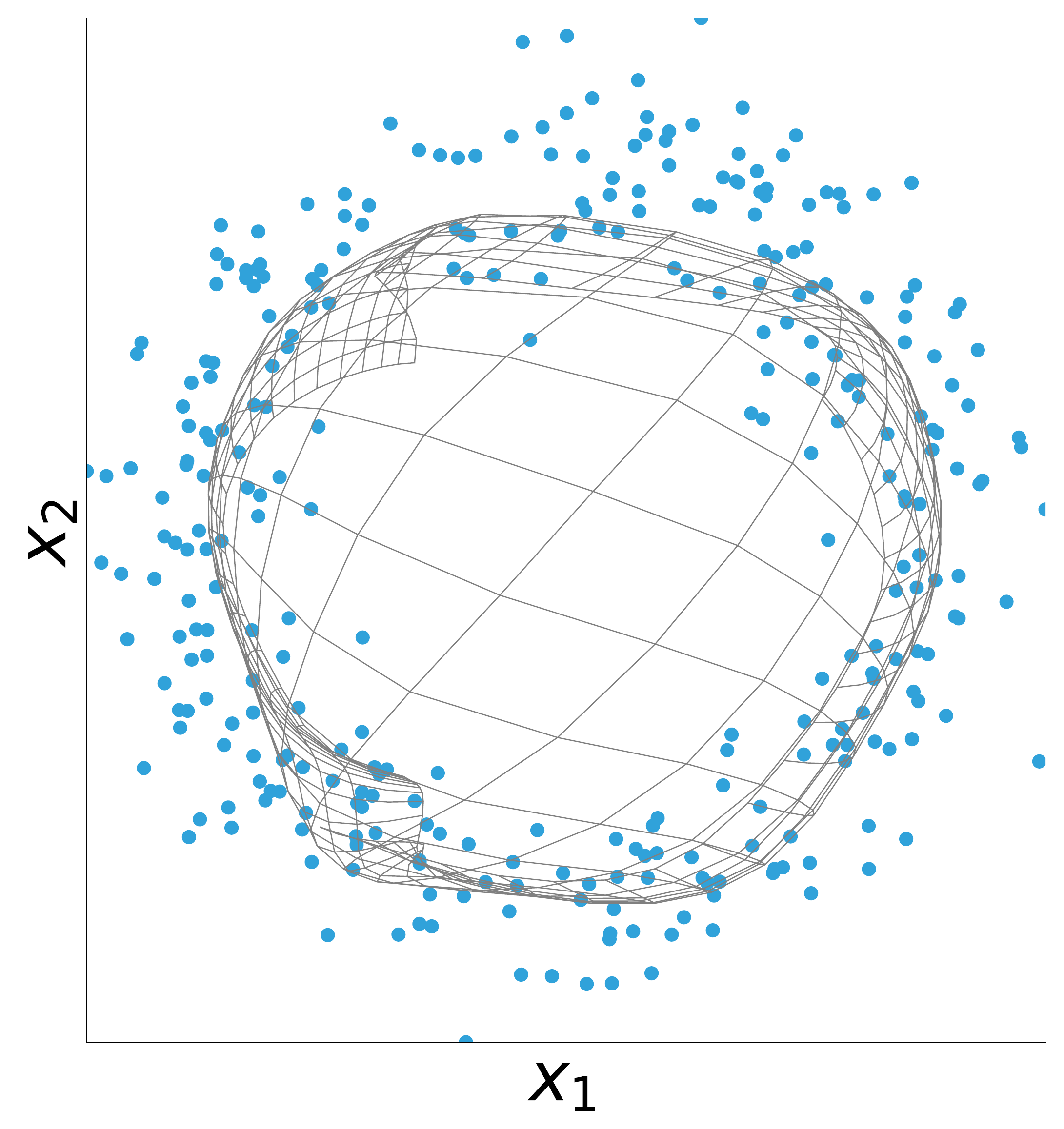}

}\subfloat[]{\includegraphics[width=0.2\textwidth]{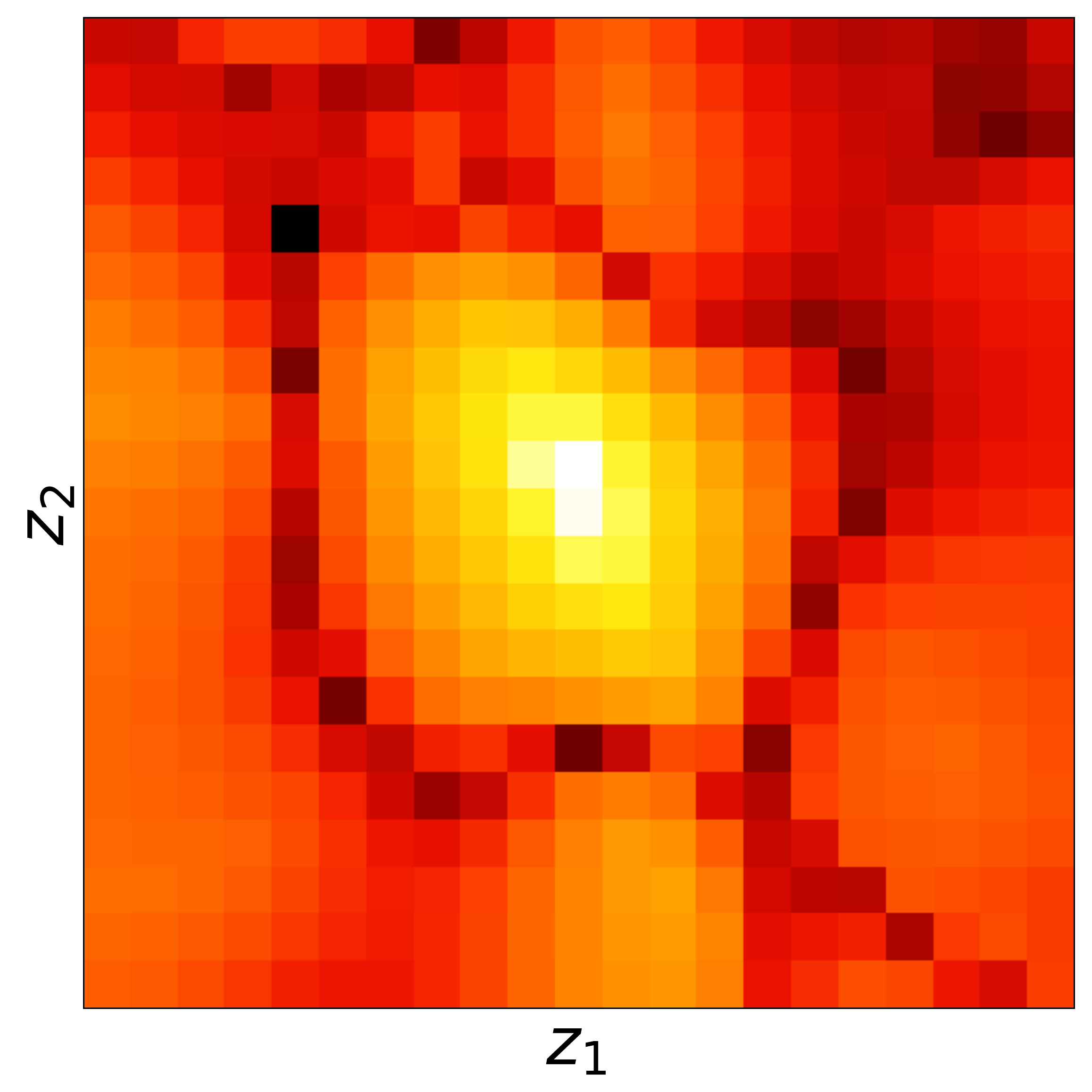}

}\subfloat[]{\includegraphics[width=0.2\textwidth]{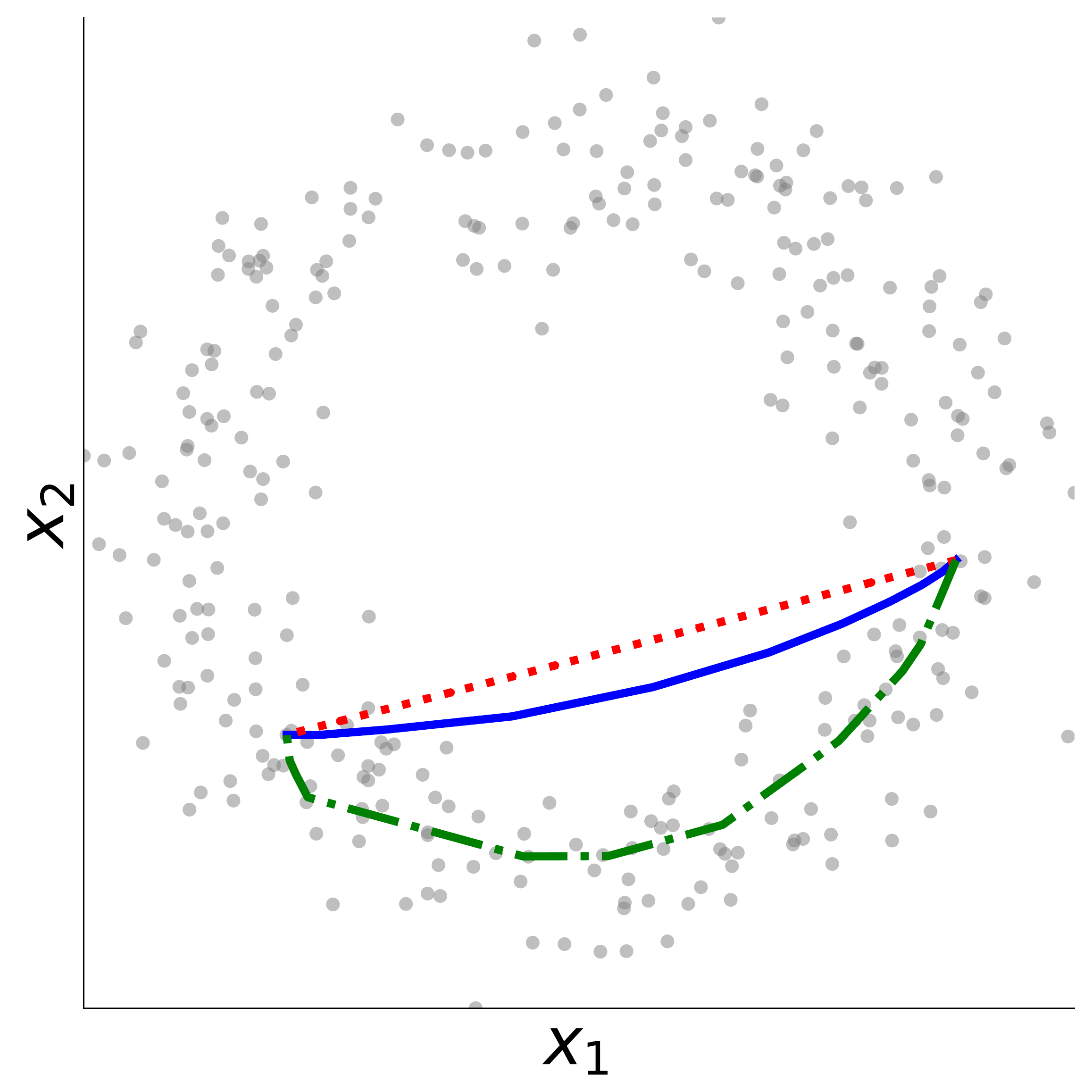}}\caption{(a) 2-D circular dataset (b) GAN (2-20-20-2) trained on the dataset.
Transformed 2-D latent space grid is shown. Note that the grid also
covers the hole in the center due to the topological constraint. (c)
$p(x)$ shown in $Z$ coordinates with heat map (i.e., $z\protect\mapsto x\protect\mapsto p(x)$).
The central area of the latent space is mapped to the hole in the
input space, thus has a low density. (d) Comparison of the interpolation
methods\protect\footnotemark -- \textbf{\emph{Blue-solid}}: linear
interpolation method found in previous literature; \textbf{\emph{Red-dotted}}:
geodesic interpolation method; \textbf{\emph{Green-dash-dotted}}:
geodesic with density regularizer. \label{fig:circular}}
\end{figure*}

\footnotetext{Note that the trained generative mapping is not injective and not an embedding at some regions. We selected two points that the pathes connecting them do not cause any problems regarding this.}

\subsection{Gradient Descent Method to Compute the Geodesic Interpolation \label{subsec:Derivation-of-the-Geodesic}}

In this section, we introduce a path-energy loss function and corresponding
gradient descent method to compute the geodesic curve between two
data points. 

As the manifold of deep generative model is defined via the latent
space and the generative mapping, curves on the manifold can be parameterized
in the latent space. The length is measured using the Riemannian metric
$g_{mn}(z)$, which is induced from the Euclidean metric of the ambient
data space $X$. The mathematical relationship can be written as $g_{mn}(z)\dot{z}^{m}\dot{z}^{n}=\delta_{kl}\dot{x}^{k}x^{l}$
using Einstein notation and the Kronecker delta, where the details
can be found in \shortcite{chen2018metrics,shao2017riemannian}.

Now, let $z(t):[a,b]\to U$ be a curve in a latent space with boundary
conditions $z(a)=z_{a}$ and $z(b)=z_{b}$. The length of $z(t)$
under metric $g$ is given as $L(z)=\int_{a}^{b}\Vert dz/dt\Vert_{g}dt=\int_{a}^{b}\left(\sqrt{\sum_{m,n}g_{mn}(z)\dot{z}^{m}\dot{z}^{n}}\right)dt$.
Instead of directly minimizing the length $L(z)$, minimizing the
energy $E(z)=\int_{a}^{b}\epsilon(z)dt=\int_{a}^{b}\left(\frac{1}{2}\sum_{m,n}g_{mn}(z)\dot{z}^{m}\dot{z}^{n}\right)dt$
gives the same result yet in an algebraically simpler form, by noting
that $L(z)=2(b-a)E(z)$ if $\Vert dz/dt\Vert_{g}=1$\footnote{Can be easily derived using Cauchy-Schwarz Inequality.}.
Thus the formulation to obtain the geodesic is reduced to

\[
z^{*}=\arg\min_{z}E(z)\quad\text{subject to}\quad\Vert\dot{z}\Vert_{g}=1.
\]
To minimize the energy $E(z)$, we compute the gradient according
to a small change $\zeta$ to the curve $z$ ($\zeta:[a,b]\to Z$
with $\zeta(a)=\zeta(b)=0$). This is given as a \emph{Gâteaux derivative}:
$\delta E(z;\zeta)=\int_{a}^{b}\left(\frac{\partial\epsilon}{\partial z}-\frac{d}{dt}\frac{\partial\epsilon}{\partial\dot{z}}\right)\zeta dt$,
which is the functional analogue of directional derivative. 

Note that the Euler-Lagrange equation is derived in a process of finding
$\delta E(z;\zeta)=0$ regardless of $\zeta$, but it is intractable
to solve since the metric $g$ described by neural network is too
complicated. We instead seek the steepest gradient change $\zeta^{*}=\arg\max_{\zeta}\delta E(z;\zeta)$,
which is derived using the Cauchy-Schwarz inequality, 
\begin{align*}
\zeta_{i}^{*} & =\frac{\partial\epsilon}{\partial z^{i}}-\frac{d}{dt}\frac{\partial\epsilon}{\partial\dot{z}^{i}}=-\sum_{k}g_{ik}(\ddot{z}^{k}+\sum_{m,n}\Gamma_{mn}^{k}\dot{z}^{m}\dot{z}^{n}),
\end{align*}
where $\Gamma_{nm}^{k}=\frac{1}{2}g^{kl}\left(\partial g_{kn}/\partial z^{m}+\partial g_{mk}/\partial z^{n}-\partial g_{mn}/\partial z^{k}\right)$
is the Christoffel Symbol. Note that $\Gamma_{nm}^{k}$ is purely
related to the local change of the coordinate bases, thus it becomes
zero in the input-space coordinates. Also, the metric is described
as Kronecker delta in input-space coordinates, so the equation is
far simplified as 
\[
\xi_{i}^{*}=-\ddot{x}_{i},
\]
where $\xi$ is a small change to the curve $x$ described in the
input-space coordinates. 

To summarize, for a curve $z(t)$ in $U\subset Z$, we transform every
point on the curve to the input space using the generative mapping,
from which the corresponding curve $x(t)$ in $M\subset X$ is obtained.
Then, we compute the acceleration $\ddot{x}$ to get the steepest
gradient descent change toward minimizing the path-length of the curve.
By iteratively updating $x\leftarrow x+\eta\ddot{x}$ ($\eta$ is
a learning rate), $x$ will converge to the shortest path. Finally,
putting $x$ back to the latent space will give the geodesic in the
latent space. More detailed algorithm can be found in \shortcite{shao2017riemannian}
and the extended version with our proposing regularizer will be described
in the next section.

\subsection{Density Regularizer}

As introduced earlier, the topological difference between the deep
generative model and the dataset brings a serious problem in the interpolation.
For example, consider a circular dataset and the geodesic between
two points in the dataset (Fig. \ref{fig:circular}). It is natural
to think that the geodesic should follow a circular path between two
points and pass through dense regions. However, as the hole in the
center is a valid region for deep generative models, the geodesic
becomes a linear path and passes through the hole (\emph{Red-dotted}
curve in Fig. \ref{fig:circular} (d)). 

In order to prevent the geodesic from finding this kind of shortcuts,
we propose to add a density regularizer term to the energy functional:
\begin{align*}
\mathcal{L}(z;\mu) & =E(z)+\mu\int_{a}^{b}\left(-\log p(z)+\frac{1}{2}\log\left|\det[g_{z}]\right|\right)dt,
\end{align*}
where $\mu>0$ is a regularization weight. The regularization term
is in fact an integrated negative log-probability-density $-\log p(x)$
along the curve, thus it keeps the curve from passing through low-density
regions. 

In Fig. \ref{fig:circular} (c) it can be seen that the density $p(x(z))$
is low in the central area of the latent-space, which is mapped to
the hole of the input data space. Introduction of the regularizer
results in an interpolation avoiding this low-density regions, while
keeping its path-length as short as possible (\emph{Green-dash-dotted}
curve in Fig. \ref{fig:circular} (d)).

\subsection{Numerical Algorithm for Geodesic Interpolation \label{subsec:Numerical-Method}}

As explained in the previous section, geodesic curve can be computed
by iteratively updating every point on the curve $z(t)$ toward the
steepest variational direction $\zeta^{*}$. In practice, we approximate
the curve with a set of points $z^{k}$. As for $\zeta^{*}$, we first
compute $\xi^{*}=-\ddot{x}$ using finite difference method, then
pull back the vectors $\xi^{*}$ to the latent space using the pseudo-inverse
of the Jacobian matrix. The contribution from the gradient of the
density regularizer is added to the obtained $\zeta^{*}$.

\begin{algorithm}[H]
\begin{algorithmic}[Geodesic Interpolation with Density Regularizer]
\Require{Initial ordered sequence $z^k$}
\State{Compute $x^k=f(z^k)$, $J^k=J(z^k)$ for $k=1,\cdots,K$}
\While{not converged}
	\State{$v^k \leftarrow \frac{x^{k+1}-x^{k}}{\Vert x^{k+1}-x^{k} \Vert}$ \qquad\qquad for $k=1,\cdots,K-1$}
	\State{$a^k \leftarrow v^k-v^{k-1}$ \qquad\qquad for $k=2,\cdots,K-1$}
	\State{$b^k \leftarrow ({J^k}^\top J^k)^{-1} {J^k}^\top a^k$ \quad for $k=2,\cdots,K-1$}
	\State{$z^k \leftarrow z^k + \eta \left(b^k - \mu \frac{\partial}{\partial z} \left(\frac{1}{2}\log \det ([g])-\log p(z) \right) \rvert_{z=z^k} \right)$}    	\State{\qquad\qquad for $k=2,\cdots,K-1$}
	\State{Compute $x^k$, $J(z^k)$ for $k=1,\cdots,K$}
\EndWhile
\end{algorithmic}

\caption{Geometric Interpolation with Density Regularizer}
\label{alg:main}

\end{algorithm}

\section{Experiments}

\subsection{Datasets, Models and Hyperparameters}

In experiments, we use two different image datasets: \emph{1) MNIST}
\shortcite{lecun1998gradient}, and \emph{2) Yale} \shortcite{belhumeur1997eigenfaces}.
Yale dataset originally consists of face images of 28 individuals,
but only a single person images (under 64 different illumination conditions)
are used to examine the illumination manifold.

We use three different deep generative models, where each of which
is paired with the datasets. \emph{1) MNIST with GAN} \shortcite{goodfellow2014generative},
\emph{2) MNIST with VAE} \shortcite{kingma2013auto}, \emph{3) Yale with
DCGAN} (Deep Convolutional GAN) \shortcite{radford2015unsupervised}. 

When running the numerical algorithm to compute the geodesic, number
of approximation points is fixed to 35. We observed that too large
number of approximation points sometimes yield an entire break-up
of the smoothness of the curve due to the approximation noise. Regularizer
weight $\mu$ is empirically determined, ranging from 0.1 \textasciitilde{}
0.003 according to each setting. 

\begin{figure*}[t]
\begin{centering}
\includegraphics[width=0.12\textwidth]{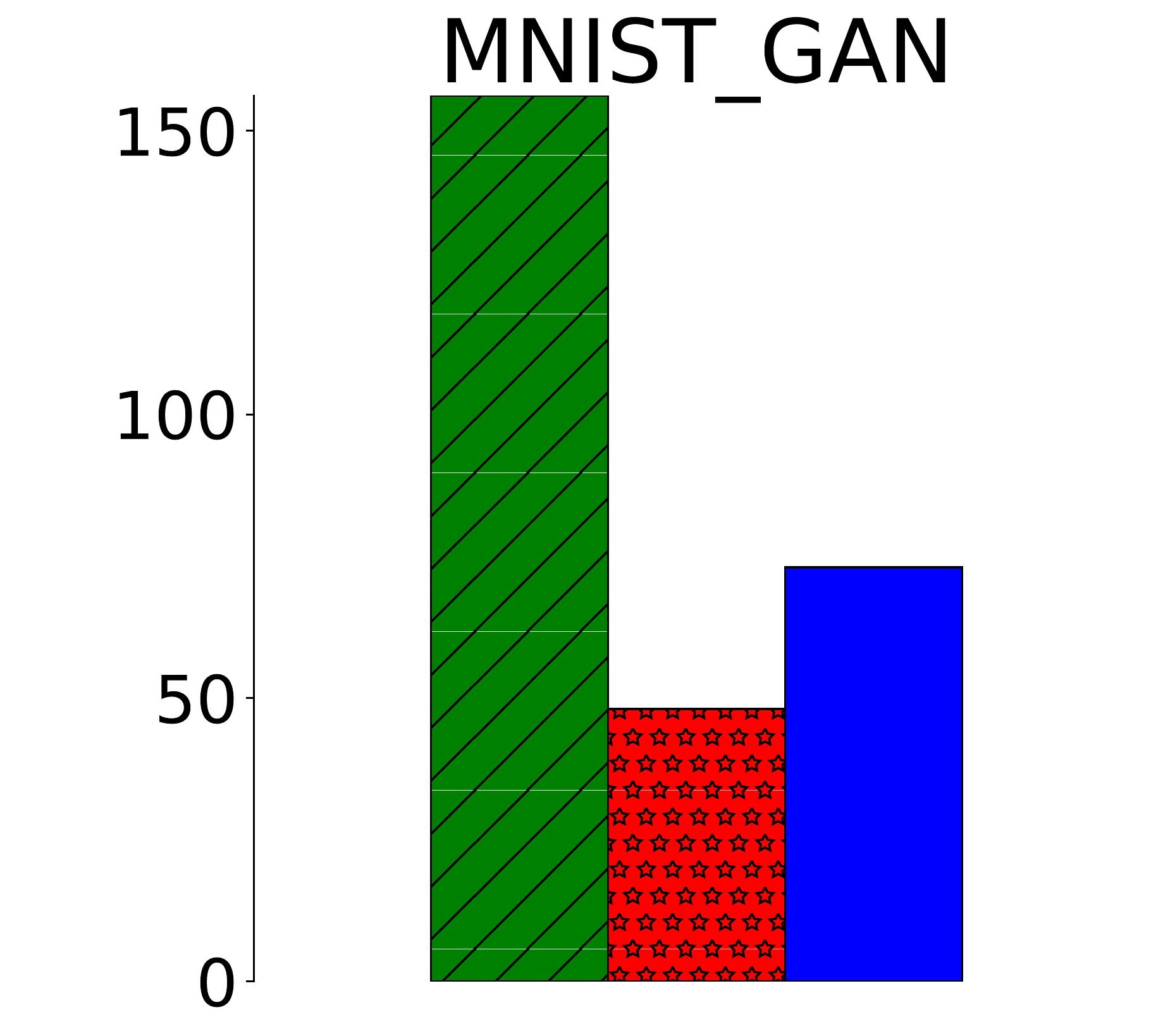}\includegraphics[width=0.64\textwidth]{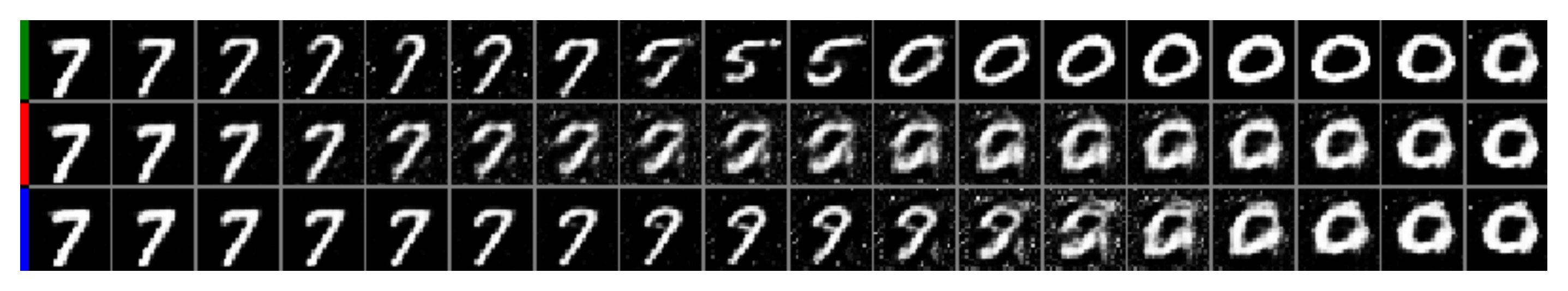}
\par\end{centering}
\begin{centering}
\vspace{-0.5em}
\par\end{centering}
\begin{centering}
\includegraphics[width=0.12\textwidth]{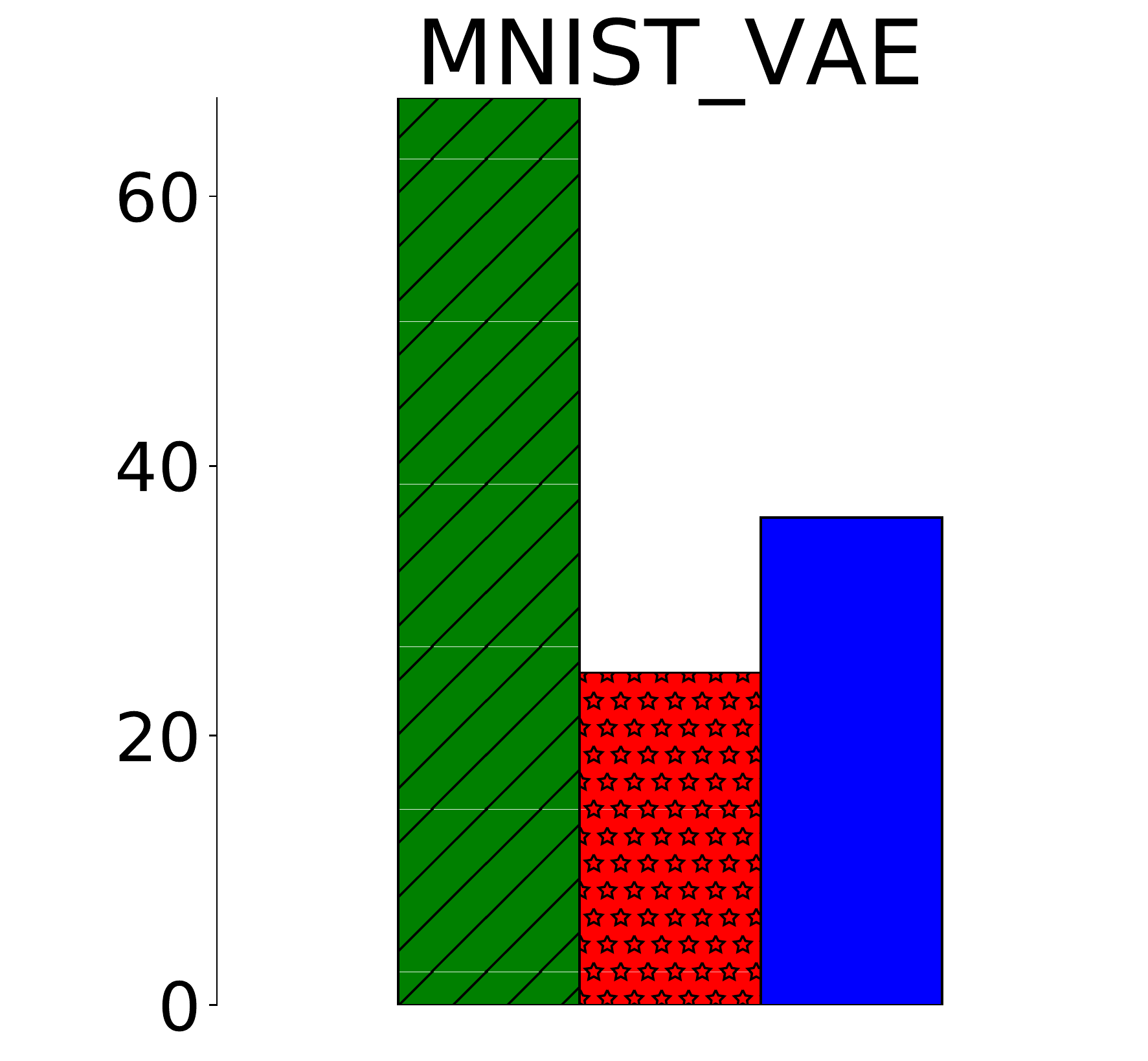}\includegraphics[width=0.64\textwidth]{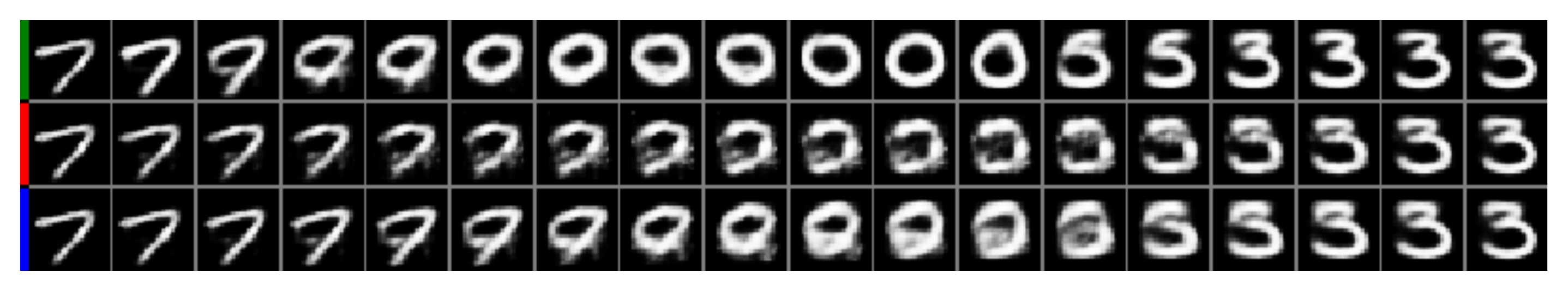}\vspace{-0.5em}
\par\end{centering}
\begin{centering}
\includegraphics[width=0.12\textwidth]{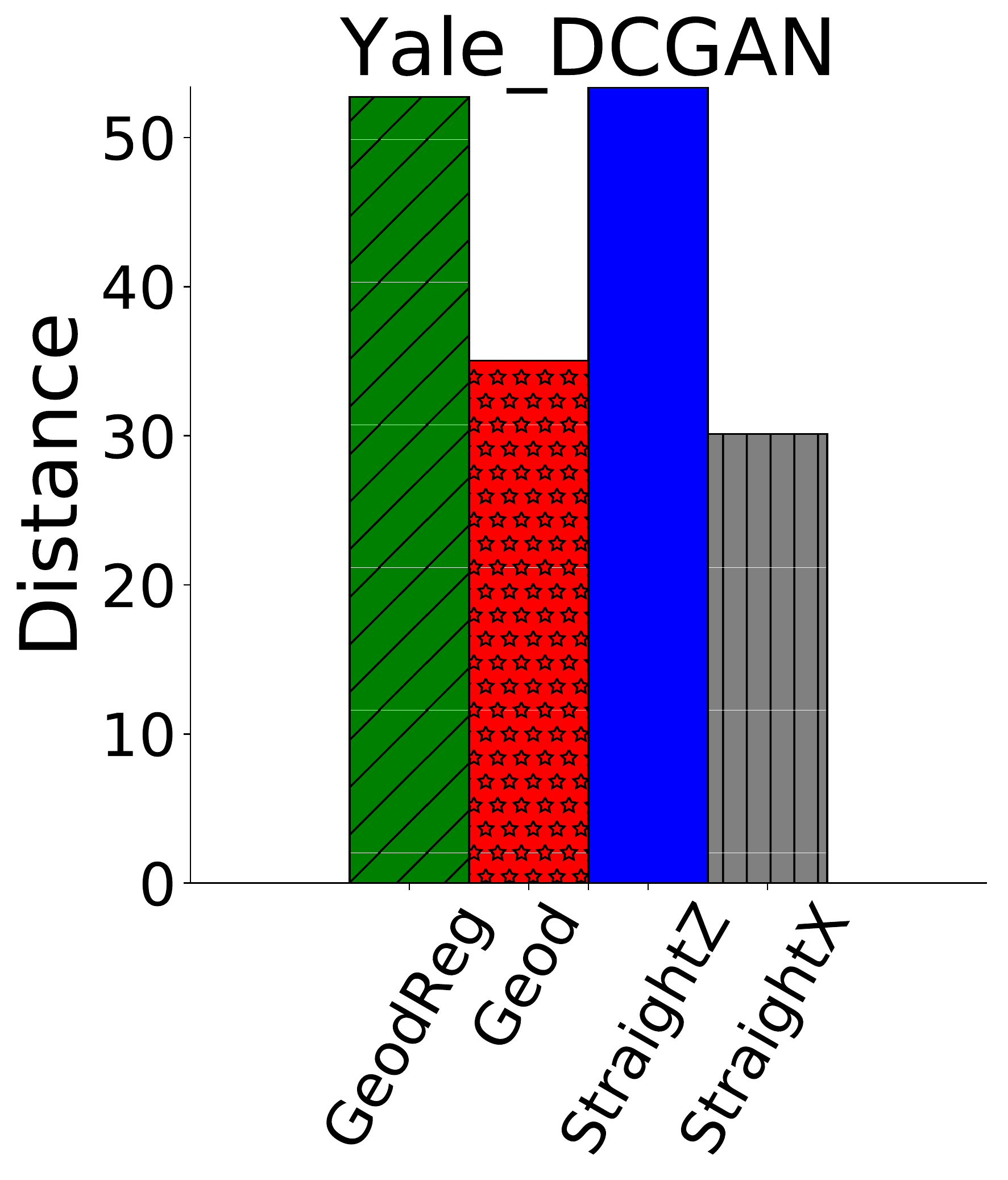}\includegraphics[width=0.64\textwidth]{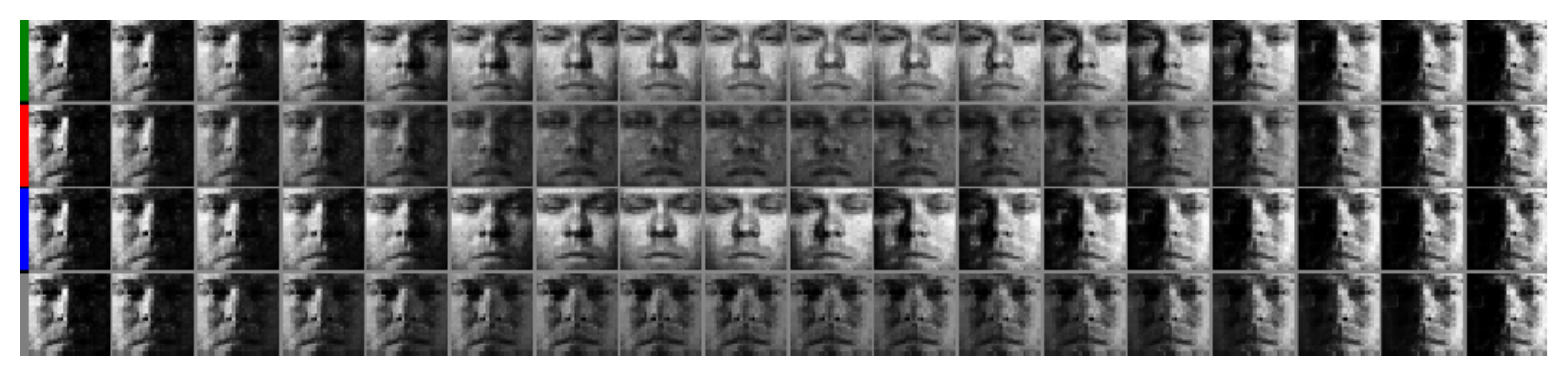}
\par\end{centering}
\begin{centering}
\includegraphics[width=0.14\textwidth]{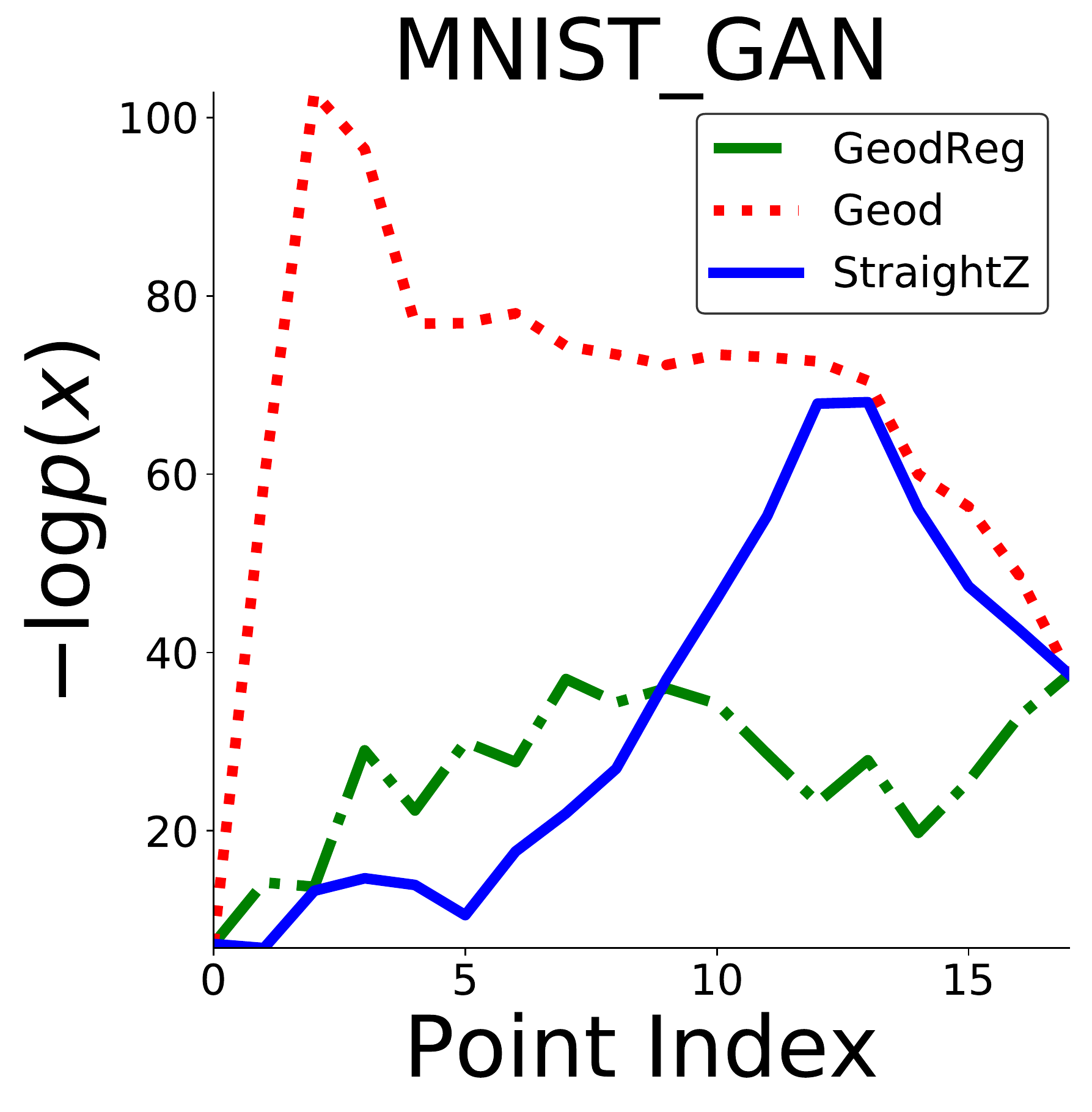}\includegraphics[width=0.14\textwidth]{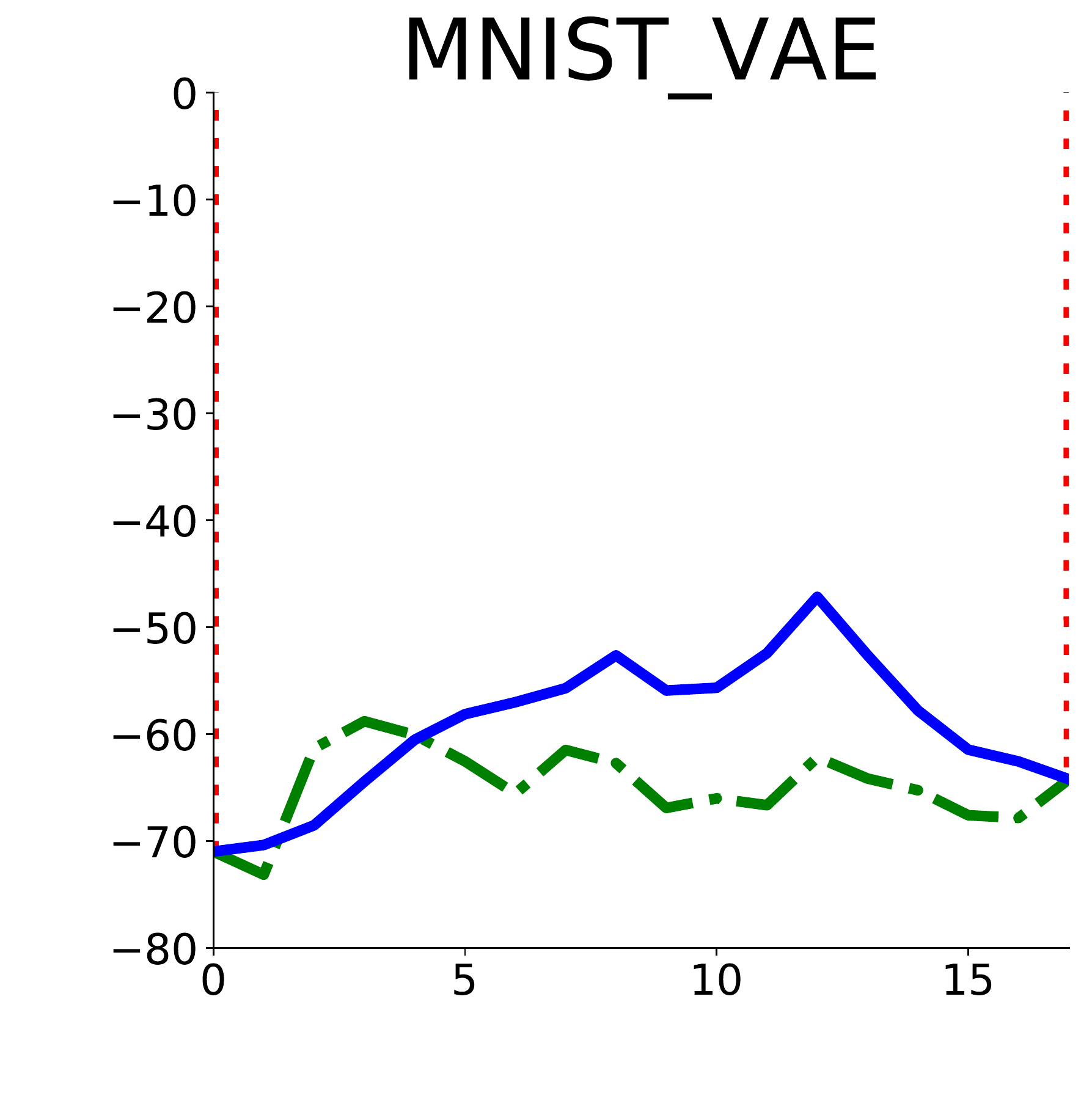}\includegraphics[width=0.14\textwidth]{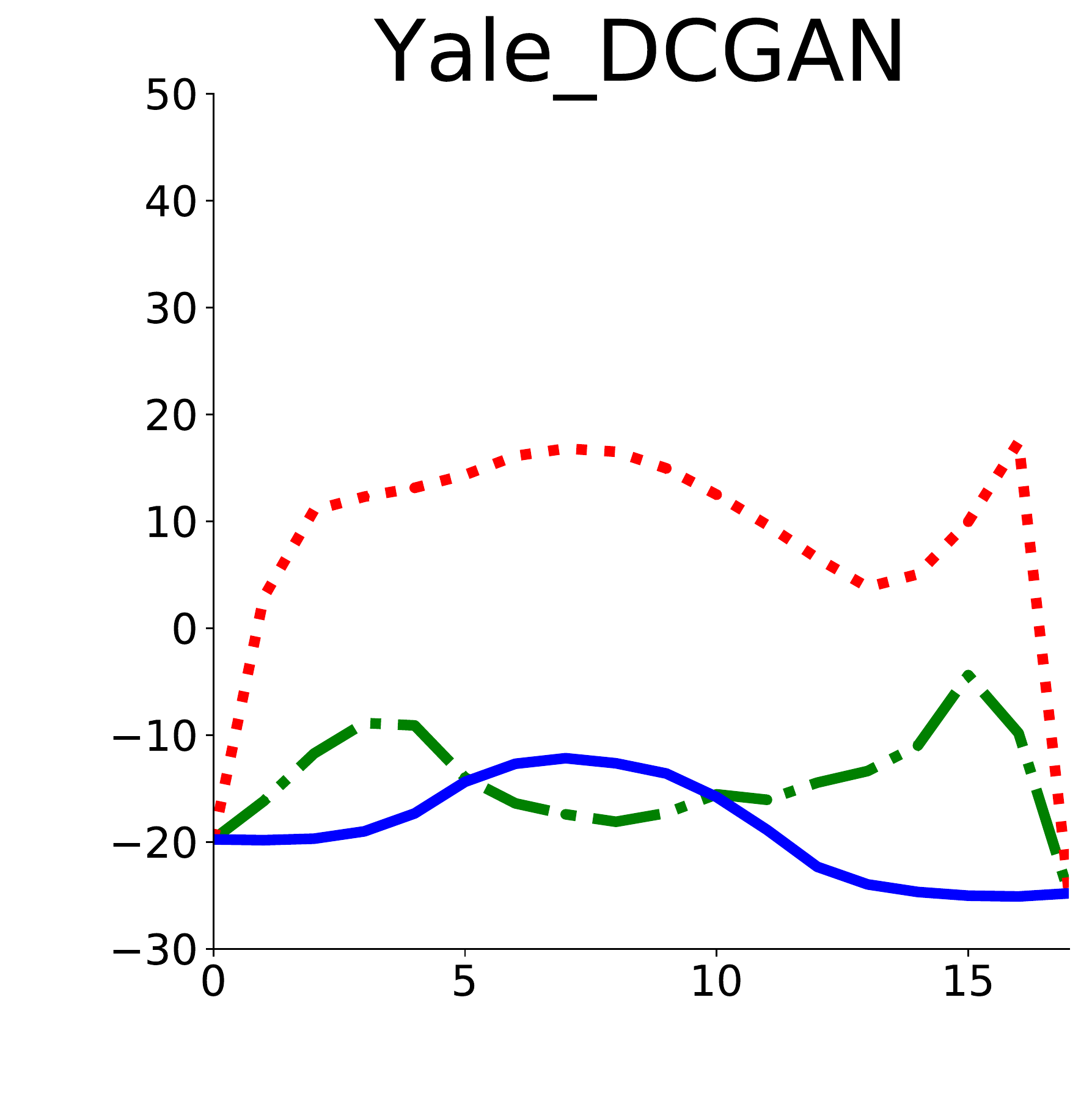}
\par\end{centering}
\caption{Interpolation results from MNIST GAN, MNIST VAE and Yale DCGAN models,
using different methods. In all plots, colors are matched according
to the interpolation methods: Geodesic with Density Regularizer (\emph{green}),
Geodesic (\emph{red}), Straight in $Z$ (\emph{blue}). Aside from
the colors, the images are organized in order of GeodReg, Geod, StraightZ,
from top to bottom. Corresponding distances are shown in the left
with \emph{striped}, \emph{dotted}, \emph{plain} bar graphs. Negative
Log-Likelihood along the interpolation points on the path are shown
at the bottom with \emph{dash-dotted}, \emph{dotted}, \emph{plain}
curves. }
\label{fig:main_result}
\end{figure*}

\subsection{Interpolations and the Effect of the Regularizer}

In Fig. \ref{fig:main_result}, we show the results from geodesic
interpolation method, with and without the density regularizer, as
well as the results from a naïve method that follow a straight line
in $Z$. Hereafter, we denote each method as \emph{GeodReg}, \emph{Geod}
and \emph{StraightZ} respectively. 

In MNIST results, it can be clearly seen that GeodReg outperforms
others in terms of quality. GeodReg shows smooth transitions along
realistic samples, whereas Geod and StraightZ contain some mottled,
non-number-like images. Geod and StraightZ do not consider the probability
density, thus cannot avoid low density regions as shown in the log-density
plot in the bottom. Geod particularly passes through much lower density
regions, but instead has the shortest length among others. Note that
these results agree with the toy example shown in Fig. \ref{fig:circular}
(d).

\subsection{Lambertian Image Manifold}

According to \shortcite{basri2003lambertian}, object images taken under
various illumination conditions form a 9-D linear manifold if the
object has Lambertian reflectance. Yale face dataset matches up to
the Lambertian condition in practice \shortcite{lee2005acquiring}, thus
we examine if our interpolation methods find linear paths on Yale
dataset.

\begin{figure*}[!t]
\begin{centering}
\includegraphics[width=0.16\textwidth]{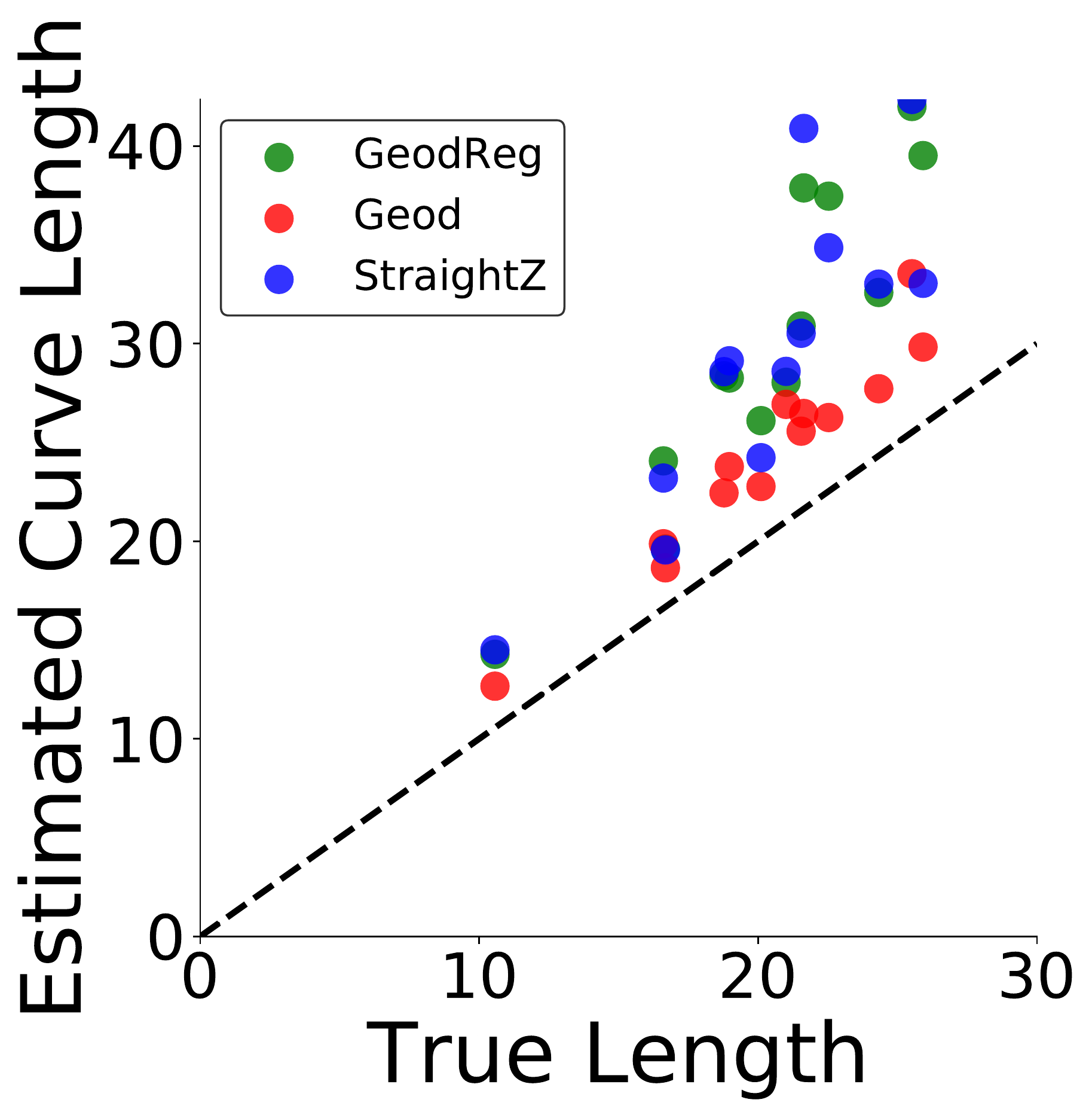}\includegraphics[width=0.4\textwidth]{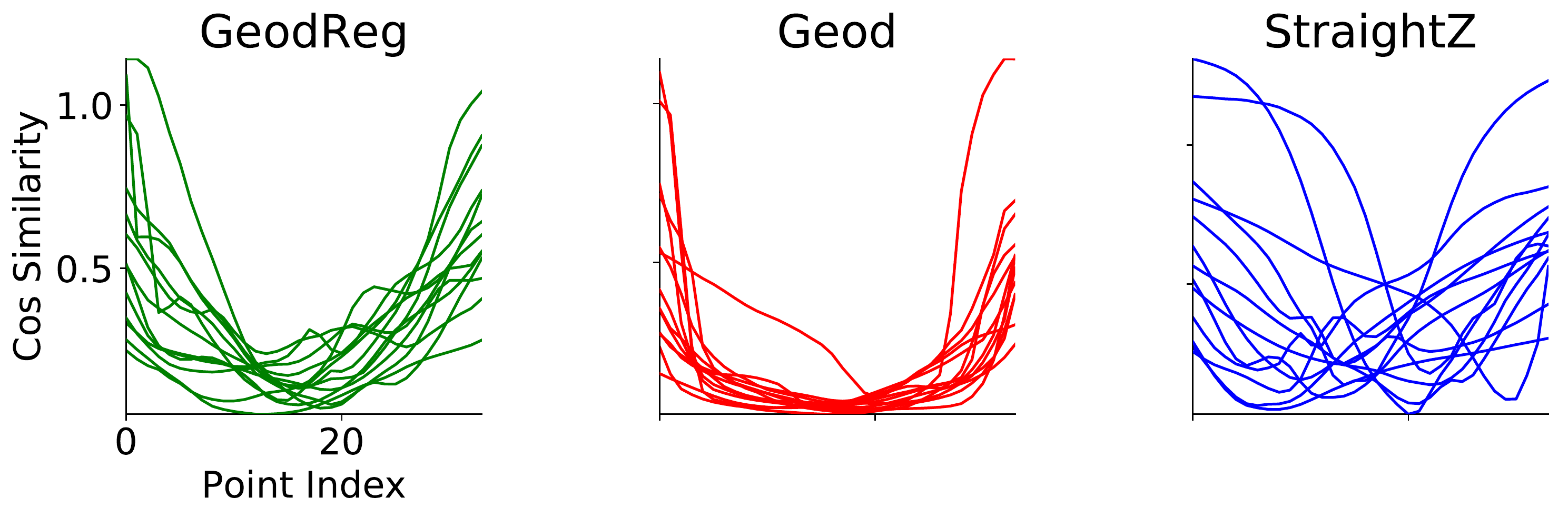}
\par\end{centering}
\centering{}\caption{Results from Yale dataset. (\emph{Left}) Lengths of the curves obtained
from different interpolation methods, compared to the ground truth
length. (\emph{Right}) Cosine dissimilarities between the tangent
vectors at each point on the curve and the direction of the true linear
path are shown. In both subfigures, Geod finds the shortest and the
straightest one, thus closest to the ground truth. Interpolations
for 13 different sample pairs are shown.}
\label{fig:lambertian}
\end{figure*}

In the left pane of the Fig. \ref{fig:lambertian}, it can be seen
that the length obtained from Geod is shortest thus closest to the
true length. In the right pane, it can be seen that the path found
by Geod is the straightest in the input space, as it has the smallest
cosine dissimilarity to the true linear path. GeodReg deviates from
the true path while seeking high density but still close to it compared
to StraightZ as seen from the right pane.

\section{Concluding Remarks}

In this paper, we have proposed a density-regularized geometric interpolation
method for deep generative models trained from non-simply-connected
datasets. The density regularizer has played a crucial role to compensate
the topological difference between the model and the data. Our method
has shown superior interpolation qualities over previous linear and
geodesic interpolation methods. \blfootnote{This work was partly supported by the Korea government (2015-0-00310, 2017-0-01772, 2018-0-00622, KEIT-10060086).}\bibliographystyle{aaai}
\bibliography{paper}

\begin{thebibliography}{}

\bibitem[\protect\citeauthoryear{Basri and Jacobs}{2003}]{basri2003lambertian}
Basri, R., and Jacobs, D.~W.
\newblock 2003.
\newblock Lambertian reflectance and linear subspaces.
\newblock {\em IEEE transactions on pattern analysis and machine intelligence}
  25(2):218--233.

\bibitem[\protect\citeauthoryear{Belhumeur, Hespanha, and
  Kriegman}{1997}]{belhumeur1997eigenfaces}
Belhumeur, P.~N.; Hespanha, J.~P.; and Kriegman, D.~J.
\newblock 1997.
\newblock Eigenfaces vs. fisherfaces: Recognition using class specific linear
  projection.
\newblock {\em IEEE Transactions on pattern analysis and machine intelligence}
  19(7):711--720.

\bibitem[\protect\citeauthoryear{Chen \bgroup et al\mbox.\egroup
  }{2018}]{chen2018metrics}
Chen, N.; Klushyn, A.; Kurle, R.; Jiang, X.; Bayer, J.; and Smagt, P.
\newblock 2018.
\newblock Metrics for deep generative models.
\newblock In {\em International Conference on Artificial Intelligence and
  Statistics},  1540--1550.

\bibitem[\protect\citeauthoryear{Goodfellow \bgroup et al\mbox.\egroup
  }{2014}]{goodfellow2014generative}
Goodfellow, I.; Pouget-Abadie, J.; Mirza, M.; Xu, B.; Warde-Farley, D.; Ozair,
  S.; Courville, A.; and Bengio, Y.
\newblock 2014.
\newblock Generative adversarial nets.
\newblock In {\em Advances in neural information processing systems},
  2672--2680.

\bibitem[\protect\citeauthoryear{Kingma and Welling}{2013}]{kingma2013auto}
Kingma, D.~P., and Welling, M.
\newblock 2013.
\newblock Auto-encoding variational bayes.
\newblock {\em arXiv preprint arXiv:1312.6114}.

\bibitem[\protect\citeauthoryear{LeCun \bgroup et al\mbox.\egroup
  }{1998}]{lecun1998gradient}
LeCun, Y.; Bottou, L.; Bengio, Y.; and Haffner, P.
\newblock 1998.
\newblock Gradient-based learning applied to document recognition.
\newblock {\em Proceedings of the IEEE} 86(11):2278--2324.

\bibitem[\protect\citeauthoryear{Lee, Ho, and
  Kriegman}{2005}]{lee2005acquiring}
Lee, K.-C.; Ho, J.; and Kriegman, D.~J.
\newblock 2005.
\newblock Acquiring linear subspaces for face recognition under variable
  lighting.
\newblock {\em IEEE Transactions on pattern analysis and machine intelligence}
  27(5):684--698.

\bibitem[\protect\citeauthoryear{Radford, Metz, and
  Chintala}{2015}]{radford2015unsupervised}
Radford, A.; Metz, L.; and Chintala, S.
\newblock 2015.
\newblock Unsupervised representation learning with deep convolutional
  generative adversarial networks.
\newblock {\em arXiv preprint arXiv:1511.06434}.

\bibitem[\protect\citeauthoryear{Shao, Kumar, and
  Fletcher}{2017}]{shao2017riemannian}
Shao, H.; Kumar, A.; and Fletcher, P.~T.
\newblock 2017.
\newblock The riemannian geometry of deep generative models.
\newblock {\em arXiv preprint arXiv:1711.08014}.

\bibitem[\protect\citeauthoryear{Tenenbaum, De~Silva, and
  Langford}{2000}]{tenenbaum2000global}
Tenenbaum, J.~B.; De~Silva, V.; and Langford, J.~C.
\newblock 2000.
\newblock A global geometric framework for nonlinear dimensionality reduction.
\newblock {\em science} 290(5500):2319--2323.

\end{thebibliography}
\newpage{}

\setcounter{page}{1}
\end{document}